\documentclass[runningheads]{llncs}
\usepackage[T1]{fontenc}
\usepackage{graphicx}
\usepackage{multirow}
\usepackage{amsfonts}
\usepackage{amsmath}
\usepackage{booktabs}
\usepackage{algorithm}
\usepackage{algorithmic}
\usepackage{makecell}
\usepackage{amsbsy}
\usepackage{multicol}
\usepackage{xcolor,colortbl}
\usepackage{kotex}
\usepackage[colorlinks=true,
            linkcolor=blue,
            urlcolor=blue,
            citecolor=blue]{hyperref}

\usepackage{bbm}
\usepackage{makecell}

\begin{document}
%

\title{\textit{MNM}: Multi-level Neuroimaging Meta-analysis with Hyperbolic Brain-Text Representations}



%

\author{Seunghun Baek
\and
Jaejin Lee
\and
Jaeyoon Sim
\and
Minjae Jeong
\and
Won Hwa Kim} 

\authorrunning{S. Baek et al.}
\titlerunning{\textit{MNM}: Multi-level Neuroimaging Meta-analysis}


\institute{
Pohang University of Science and Technology, Pohang, South Korea \\
\email{\{habaek4, jlee4923, simjy98, minjaetidtid, wonhwa\}@postech.ac.kr}
}

\maketitle              




\begin{abstract}
Various neuroimaging studies suffer from small sample size problem which often limit their reliability. 
Meta-analysis addresses this challenge by 
aggregating findings from different studies to identify consistent patterns of brain activity. 
However, traditional approaches based on keyword retrieval or linear mappings often overlook the rich hierarchical structure in the brain. 
In this work, we propose a novel framework that leverages hyperbolic geometry to bridge the gap between neuroscience literature and brain activation maps. 
By embedding text from research articles and corresponding brain images into a shared hyperbolic space via the
Lorentz model, our method captures both semantic similarity and hierarchical organization inherent in neuroimaging data.
In the hyperbolic space, our method performs multi-level neuroimaging meta-analysis (MNM) by 
1) aligning brain and text embeddings for semantic correspondence,
2) guiding hierarchy between text and brain activations,
and 3) 
preserving the hierarchical relationships within brain activation patterns. 
Experimental results demonstrate that our model outperforms baselines, offering a robust and interpretable paradigm of multi-level neuroimaging meta-analysis via hyperbolic brain-text representation.

\keywords{Neuroimaging Meta-analysis \and Hyperbolic Representation.}

\end{abstract}

\section{Introduction}

Recent research in neuroscience is rapidly expanding our understanding of brain function and structure,
uncovering diverse aspects of neural organization across various regions \cite{sim2024multi,kim2014multi}.
However, neuroimaging studies are often underestimated 
due to 
small sample sizes of their experiments~\cite{reliability1,reliability2}. 
Brain meta-analysis tackles these challenges by integrating neuroscientific descriptions (i.e., body text) with corresponding brain activation maps from multiple studies to identify consistent patterns in the brain.
Traditional approaches~\cite{brainmap,neurosynth,neuroquery} have relied on predefined keywords and regression models to discover these patterns by predicting brain activation maps from textual descriptions.
More recent approaches utilize large language models (LLMs) to handle longer sequences \cite{text2brain}, 
and employ contrastive learning to map textual descriptions and activation coordinates into a joint representation space \cite{neurocontext2} for text-to-brain activation prediction. 

\begin{figure}[t!]
\centering
\includegraphics[width=0.9\textwidth]{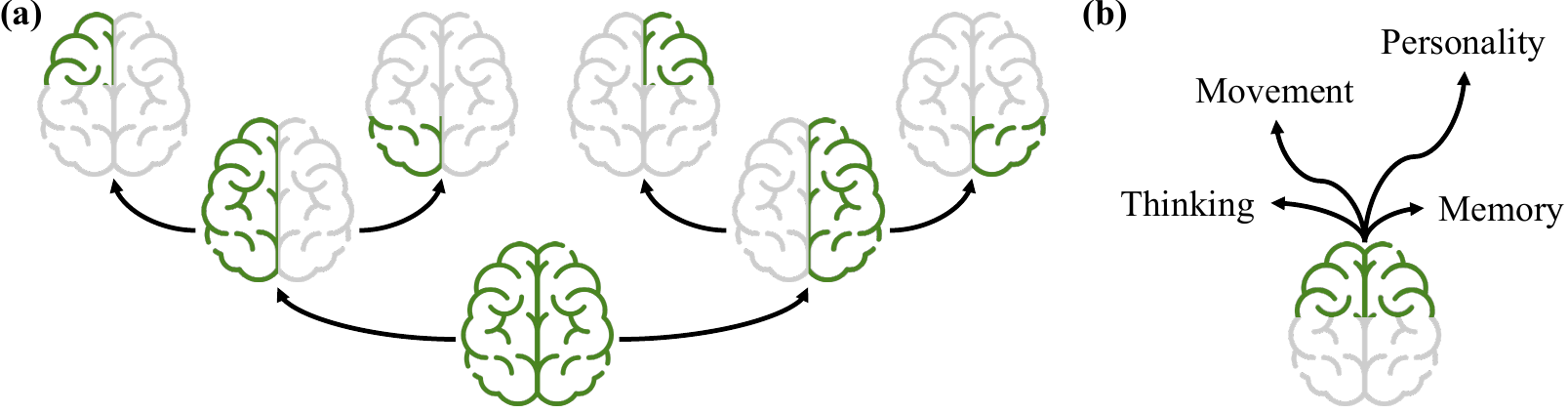}
    \caption{\footnotesize 
    \textbf{Motivation of MNM.}
    (a) The brain exhibits a spatial hierarchy, where broad regions can be further subdivided into more specific regions.
    (b) Each brain region can correspond to various articles covering diverse topics, such as functionalities.
    }
    \label{fig:hierarchy_concept}
\end{figure}

Notice that the human brain exhibits a naturally nested organization, 
where broad regions are gradually subdivided into more specialized areas as in Fig.~\ref{fig:hierarchy_concept}(a).
Capturing these {\em multi-level relationships} is essential for robustly linking textual descriptions from a coarse activation (i.e, involved hemisphere) to fine details (i.e., the most relevant region of interests (ROIs)).
Also, as depicted in Fig.~\ref{fig:hierarchy_concept}(b),
each brain region is defined by its unique functionality 
and can be linked to numerous articles that explore diverse topics related to these functions. 
Such insights on hierarchy are not included in previous approaches, where they map brain activation and text pairs to the identical point in the Euclidean space \cite{neurocontext2}. 

To better jointly represent brain activation and text pairs, we 
rely on 
hyperbolic spaces with exponentially expanding geometry, which are 
suited for representing tree-like structures.
Therefore, we propose to embed 
the corresponding pairs 
into a shared hyperbolic space, together with an angle-based contrastive learning to capture mutual context between brain activation and text. 
Our 
framework, {\bf M}ulti-level {\bf N}euroimaging {\bf M}eta-analysis with Hyperbolic Brain-Text Representations (MNM), preserves these inherent hierarchical relationships and overcomes the limitations of the Euclidean method, which lead to the following 
\textbf{major contributions:} 
\textbf{1)} We introduce a hyperbolic embedding framework for neuroimaging meta-analysis that naturally captures semantic associations between brain and text and brain's hierarchical structure.
\textbf{2)} 
We propose a brain structural hierarchy guidance such that brain structural hierarchy is preserved in the hyperbolic space. 
\textbf{3)} MNM is validated through extensive experiments on neuroimaging meta-analysis, such as cross-modal retrieval and brain activation prediction, demonstrating enhanced semantic alignment and interpretability in the same 
setting in \cite{neurocontext2}. 
Cross-modal retrieval and brain activation map prediction demonstrate that our approach not only improves semantic alignment between brain and text modalities but also preserves the multi-level organization of brain regions. 
Hierarchical structure analyses and ablation studies further confirm that incorporating both brain structural and brain–text hierarchies leads to more interpretable analyses and enhanced retrieval performance.

\textbf{Related Works on Neuroimaging Meta-Analysis.}
Previous approaches for brain meta-analysis have transitioned from manual to automated methods. 
BrainMap~\cite{brainmap} manually curated neuroimaging articles, 
a strategy that struggles to scale with the growing size of literature.  
NeuroSynth~\cite{neurosynth} and NeuroQuery~\cite{neuroquery} automate the extraction of brain activation coordinates using fixed keywords and statistical techniques, 
yet they are limited by bag-of-words representations that fail to capture complex semantics. 
Text2Brain~\cite{text2brain} addresses this by using SciBERT~\cite{scibert} embeddings and convolutional neural networks to generate 3D activation maps, 
though it faces challenges with longer texts. 
Recently, NeuroConText~\cite{neurocontext2} employs a contrastive framework that leverages advanced language models such as Mistral-7B~\cite{mistral7b} alongside the Dictionary of Functional Mode (DiFuMo) atlas~\cite{difumo} to 
enhance the alignment between paired modalities.
While we adopt identically processed
dataset from \cite{neurocontext2},
MNM outperforms it by leveraging hyperbolic geometry to capture the brain's intrinsic hierarchical structure,
thereby achieving superior semantic alignment and more flexible multi-level meta-analysis.

\section{\textit{Preliminary}: Hyperbolic Spaces and Lorentz Model}
\label{sec:hyperbolic}

Unlike 
Euclidean 
space,
hyperbolic space expands exponentially under constant negative curvature, making it highly effective for 
representing hierarchical structures. 
This 
enables capturing multi-level traits in both brain structure and text. 

\noindent\textbf{Lorentz model.} 
Lorentz model $\mathbb{L}^n$ expresses the $n$-dimensional hyperbolic space 
with curvature $c$. It is defined in a $(n+1)$-dimensional Euclidean space $\mathbb{R}^{n+1}$ as 
$\mathbb{L}^n = \{ \textbf{x} \in \mathbb{R}^{n+1} \mid {\langle \textbf{x}, \textbf{x} \rangle}_\mathbb{L} = -1/c, x_0 > 0 \}$, 
where Lorentzian inner product 
is defined as $    {\langle \textbf{x}, \textbf{y} \rangle}_\mathbb{L} = -x_0y_0 + \sum_{i=1}^d{x_iy_i}$. 
Following \cite{minkowski},
$\mathbf{x}\in\mathbb{L}^{n}$ is expressed as $[x_\mathrm{time}\in\mathbb{R}, \mathbf{x}_\mathrm{space}\in\mathbb{R}^n]$.
Then, $x_\text{time}$ is determined 
as: 
\begin{equation}
\small
    x_\text{time}=\sqrt{1/c+\sum_{i=1}^d{x_i^2}}.
\end{equation}


\noindent\textbf{Geodesics.}
The shortest distance between two points $\mathbf{x}$, $\mathbf{y}$ 
on the hyperboloid is called Lorentzian distance $d_\mathbb{L}(\mathbf{x},\mathbf{y})$ and given by: 
\begin{equation}
\small
d_\mathbb{L}(\mathbf{x},\mathbf{y}) = \sqrt{1/c}\text{ }\cosh^{-1}(-c\langle \textbf{x}, \textbf{y} \rangle_{\mathbb{L}}).
\end{equation}

\noindent\textbf{Tangent space.} 
Tangent space at a point \textbf{u} $\in$ $\mathbb{L}^n$ is an Euclidean space spanned by a set of vectors orthogonal to \textbf{u}.
For $\textbf{z}\in\mathbb{R}^{n+1}$ 
in the tangent space, \textit{exponential map} $\mathbf{Exp_u(z)}$ transforms $\textbf{z}$ into $\mathbf{x}\in\mathbb{L}^n$ in the hyperbolic space as
\begin{equation}
\small
    \mathbf{x} = \mathbf{Exp_u(z)} = \cosh{(\sqrt{c}\sqrt{\langle \textbf{z}, \textbf{z} \rangle}_{\mathbb{L}})}\mathbf{u}+\frac{\sinh(\sqrt{c}\sqrt{\langle \textbf{z}, \textbf{z} \rangle}_{\mathbb{L}})}{\sqrt{c}\sqrt{\langle \textbf{z}, \textbf{z} \rangle}_{\mathbb{L}}}\mathbf{z}.
\end{equation}
With the hyperboloid origin $\mathbf{O}= [1/c, \mathbf{0}]$ introduced in \cite{meru}, 
$\mathbf{x}_\mathrm{space}$ is defined as
\begin{equation}    
\label{eq:expmap}
\small
    \mathbf{x}_\mathrm{space} = \mathbf{Exp}_{[1/c, 0]}\mathbf{(z)} = \frac{\sinh(\sqrt{c}\sqrt{\langle \textbf{z}, \textbf{z} \rangle}_{\mathbb{L}})}{\sqrt{c}\sqrt{\langle \textbf{z}, \textbf{z} \rangle}_{\mathbb{L}}}\mathbf{z}
\end{equation}

\noindent\textbf{Centroid.} 
In the hyperbolic space, the \textit{centroid} of $N$ points in Lorentz space can be computed in the Klein space \( \mathbb{K}^n \).
Conversion between two spaces are

\begin{equation}
\small
T_{\mathbb{L} \to \mathbb{K}}(\mathbf{x}) = \frac{\mathbf{x}_{\mathrm{space}}}{x_{\mathrm{time}}}, 
\quad 
T_{\mathbb{K} \to \mathbb{L}}(\mathbf{k}) = \frac{[1, \mathbf{k}]}{\sqrt{c(1 - \|\mathbf{k}\|^2)}},
\end{equation}
where $\mathbf{k} \in \mathbb{R}^n$ is a point in  \( \mathbb{K}^n \). 
Lorentzian centroid is then defined as
\begin{equation}    
\small
\label{eq:centroid}
\text{centroid}_\mathbb{L} \left( \{\mathbf{x}_j\}_{j=1}^{N} \right) = 
T_{\mathbb{K} \to \mathbb{L}} \left( \frac{\sum_{j=1}^{N} \gamma_j T_{\mathbb{L} \to \mathbb{K}}(\mathbf{x}_j)}{\sum_{j=1}^{N} \gamma_j} \right),
\end{equation}
where the Lorentz factors $\gamma_j\in\mathbb{R}$ are defined as 
$1/\sqrt{1 - c\|T_{\mathbb{L} \to \mathbb{K}}(\mathbf{x}_j)\|^2}$.


\section{Hyperbolic Brain-Text Representation}



\begin{figure}[t!]
\centering
\includegraphics[width=0.94\textwidth]{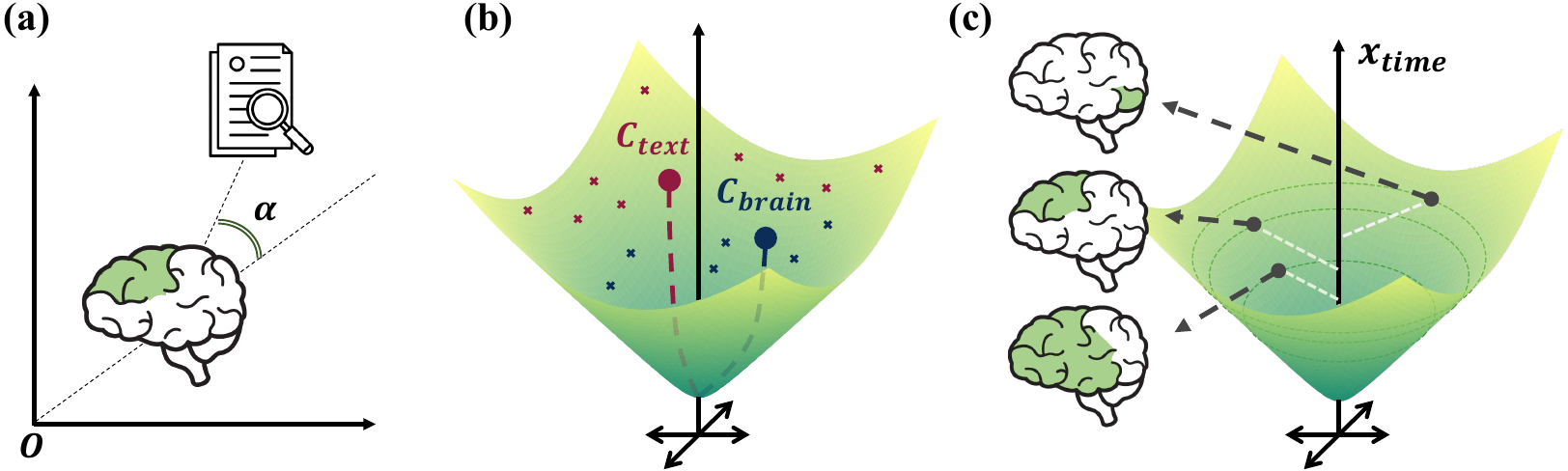}\\
    \caption{\footnotesize 
    \textbf{Concept of Our Objective Functions.}  
    (a) The angle-based contrastive loss minimizes the exterior angle between a text embedding and its corresponding brain activation embedding. 
    (b) The centroid loss encourages the centroid of brain embeddings $\mathbf{C}_\mathrm{brain}$ to be positioned closer to the 
    hyperboloid origin
    than the centroid of text embeddings $\mathbf{C}_\mathrm{text}$.  
    (c) The hierarchical loss aligns brain embeddings along the hyperbolic time axis 
    by 
    the number of activated regions
    for brain structural hierarchy.
    }
    \label{fig:loss_concept}
\end{figure}


Given a neuroscience article indexed by $i$, let $(b_i, t_i)$ denotes a pair of brain activation data and LLM-representation of its corresponding neuroscientific text.  
For neuroimaging meta-analysis, which seeks to identify the most relevant brain regions from a given textual description or retrieve corresponding brain activations,  
our goal is to embed $b_i$ and $t_i$ into a shared 
space while preserving both the brain-text hierarchy as well as the structural hierarchy of the brain.

\subsubsection{Angle-based Contrastive Learning.} 
Unlike contrastive learning~\cite{meru} that minimizes the spatial distances in a hyperbolic space,
we adopt an angle-based similarity measure~\cite{atmg} since regularizing geodesic distance weakens the hierarchical dependency.
We employ two parallel encoders: a brain activation encoder $\mathcal{E}_\text{brain}$ and a text encoder $\mathcal{E}_\text{text}$.
After mapping the $b_i$ and $t_i$ into their respective Euclidean latent spaces, 
we project each of
these onto $d$-dimensional hyperbolic space $\mathbb{L}^d$ 
via Eq.~\eqref{eq:expmap} as 
$\mathbf{z}^\mathrm{text}_i=\mathcal{E}_\mathrm{text}(t_i)$ and 
$\mathbf{z}^\mathrm{brain}_i=\mathcal{E}_\mathrm{brain}(b_i)$.
Since each brain region can be characterized by different functionalities as depicted in Fig.~\ref{fig:hierarchy_concept}(b),
we assume that brain activations have higher hierarchy (i.e., broader context) than textual descriptions.
Denoting $z_{i,\mathrm{time}}^{\mathrm{\cdot}}$ and $\mathbf{z}_{i,\mathrm{space}}^{\cdot}$ as the time and space component of $\mathbf{z}_{i}^{\cdot}$,
exterior angle
of $\mathbf{z}_{i}^{\mathrm{text}}$ from $\mathbf{z}_{i}^{\mathrm{brain}}$
is then defined as
\begin{equation}
\small
    \mathrm{ext}(\mathbf{z}_\mathrm{brain}, \mathbf{z}_\mathrm{text}) = \cos^{-1} \left( 
\frac{
    z_{i,\mathrm{time}}^\mathrm{text} + z_{i,\mathrm{time}}^\mathrm{brain} \cdot c \langle \mathbf{z}^\mathrm{brain}, \mathbf{z}^\mathrm{text} \rangle_\mathbb{L}
}{
    \|\mathbf{z}_{i,\mathrm{space}}^\mathrm{brain}\| \sqrt{(c \langle \mathbf{z}^\mathrm{brain}, \mathbf{z}^\mathrm{text} \rangle_\mathbb{L})^2 - 1}
}
\right).
\end{equation}
As exterior angle is defined in $[-\pi,\pi]$,
smaller absolute angle indicates higher correspondence.
To increase similarity between corresponding pair and discrepancies between non-corresponding pairs,
we utilize InfoNCE loss~\cite{infonce} to define our angle-based loss by minimizing the absolute angles of positive pairs as 
\begin{equation}
\small
    \mathcal{L}_{\text{angle}} = - \frac{1}{N} \sum_{i=1}^{N} \log \frac{\exp\left(-|\text{ext}(\textbf{z}_i^{\text{img}}, \textbf{z}_i^{\text{text}})| / \tau \right)}{\sum_{j=1}^{N} \exp\left(-|\text{ext}(\textbf{z}_i^{\text{img}}, \textbf{z}_j^{\text{text}})| / \tau \right)}.
\end{equation}

\subsubsection{Centroid Regularization.} 
To reflect the hierarchy between the brain activations and text,
$\mathbf{z}^\mathrm{brain}$ needs to be positioned closer to the hyperboloid origin $\mathbf{O}$ than $\mathbf{z}^\mathrm{text}$ as a smaller distance from $\mathbf{O}$ corresponds to a higher hierarchy.
However, $\mathcal{L}_\mathrm{angle}$ alone does not have explicit control over the spatial position of the embeddings.
To explicitly enforce this ordering, 
we first compute the centroids of the brain activation embeddings and the article text embeddings, 
denoted by $\mathbf{c}_\mathrm{brain}$ and $\mathbf{c}_\mathrm{text}$ each, as described in Eq.~\eqref{eq:centroid}. 
We then impose a constraint that guides these centroids toward predetermined target distances respectively as
\begin{equation}
\small
    \mathcal{L}_{\mathrm{cent}}=\left\lVert d_\mathbb{L}(\mathbf{O}, \mathbf{c}_\mathrm{text})-\frac{1}{\sqrt{c}}\cosh^{-1}(cp) \right\rVert + \left\lVert d_\mathbb{L}(\mathbf{O}, \mathbf{c}_\mathrm{img}) - \frac{1}{\sqrt{c}} \cosh^{-1}(cq) \right\rVert,
\end{equation}
where $p,q\in\mathbb{R}^+$ are hyperparameters satisfying $p>q$ 
and $||\cdot||$ is the Euclidean norm.
This condition ensures that the `broader' brain activation embeddings are positioned closer to the hyperboloid origin than the `detailed' text embeddings.

\subsubsection{Brain Structural Hierarchy Guidance.}
Brain activation map exhibits varying levels of specificity from global regions (e.g., left hemisphere) to local regions (e.g., superior temporal gyrus) as illustrated in Fig.~\ref{fig:hierarchy_concept}(a). 
To refine this structural hierarchy among brain embeddings, 
we introduce a brain structural hierarchy guidance based on the number of the activation regions.
Specifically, we set $R_i$ as how many ROIs in $b_i$ have values greater than predefined threshold $\delta$.
The brain structural hierarchy guidance
between two brain embeddings
is defined as
\begin{equation}
\small
    \mathcal{L}_\mathrm{hier} = \frac{1}{N^2} \sum_{i=1}^N \sum_{j=1}^N \mathbbm{1}_{ij}(R_i - R_j) \cdot \max(\log(z_{i,\mathrm{time}}^{\mathrm{brain}} / z_{j,\mathrm{time}}^{\mathrm{brain}}), 0),
\end{equation}
where $\mathbbm{1}_{ij}$ is an indicator function that equals 1 if $R_i>R_j$.
This 
penalizes 
when
a broader brain activation (i.e., a more general pattern) is 
wrongly 
placed at a lower hierarchical level (i.e., with a larger time value) than a 
localized activation.


\subsubsection{Joint Optimization.}
The final training loss is a weighted combination as:
\begin{equation}
\small
    \mathcal{L} = \mathcal{L}_\mathrm{angle} + \lambda_{1}\mathcal{L}_\mathrm{cent} + \lambda_{2}\mathcal{L}_\mathrm{hier},
\end{equation}
where $\lambda_1,\lambda_2\in\mathbb{R}^+$ are hyperparameters that balance the contributions of centroid regularization and brain hierarchical loss. 
By jointly minimizing this objective, the model learns to capture both the semantic alignment and the hierarchical structure necessary for robust neuroimaging meta-analysis.

\section{Experimental Settings}

\subsubsection{Data Preparation.}
Our work follows the setting established by NeuroConText~\cite{neurocontext2}, 
which processed 20,674 neuroscientific articles from PubMed along with their corresponding brain activation peak coordinates.
With a pretrained LLM (i.e., Mistral-7B),
text representation $t_i$ is derived from 
textual descriptions (e.g., abstract or body).
A corresponding discrete brain activation peak coordinate is converted into continuous brain activation voxel maps via Kernel Density Estimation (KDE),
and subsequently projected onto a 512-dimensional coefficients using the Dictionary of Functional Modes (DiFuMo) atlas as $b_i$.



\subsubsection{Implementation Details.}
For $\mathcal{E}_\mathrm{text}$ and $\mathcal{E}_\mathrm{brain}$,
we use a two-layer and three-layer residual MLP, respectively, followed by layer normalization.
We train the model for 200 epochs using the AdamW optimizer~\cite{adamw} with a learning rate of $1 \times 10^{-4}$, a batch size of 4096, and a weight decay of 0.05.  
We set $\lambda_1 = 0.5$, $\lambda_2 = 30$, 
$\delta = 5$, 
$p = 2.0$, and $q = 0.5$.  
We use pretrained weights for Text2Brain~\cite{text2brain} and NeuroQuery~\cite{neuroquery}, 
while NeuroConText~\cite{neurocontext2} is trained under the same settings as our model.
\textit{The code and dataset will be publicly available.}

\begin{table*}[b]
\caption{\footnotesize Comparison of Brain-Text Cross-Modal Retrieval Performance using two types of textual descriptions. The best performance is indicated in \textbf{bold}. 
}
\centering
\renewcommand{\arraystretch}{1.0}
\renewcommand{\tabcolsep}{0.18cm}
\scalebox{0.645}{
\begin{tabular}{cl||ccc|ccc}
    \Xhline{4\arrayrulewidth}

    \multicolumn{2}{c||}{\textbf{Retrieval}} & \multicolumn{3}{c|}{\textbf{Text$\rightarrow$Brain}} & \multicolumn{3}{c}{\textbf{Brain$\rightarrow$Text}} \\
    \hline

    \multicolumn{2}{c||}{\textbf{Metric [\%]}}
    & \textbf{Recall@5} & \textbf{Recall@10} & \textbf{Recall@100} & \textbf{Recall@5} & \textbf{Recall@10} & \textbf{Recall@100} \\
    \Xhline{2.5\arrayrulewidth}

    \multirow{6}{*}{\rotatebox[origin=c]{90}{Abstract}} & Text2Brain~\cite{text2brain}
    & 0.247$\pm$0.026 & 0.469$\pm$0.022 & 4.837$\pm$0.075 & - & - & - \\
    
    & NeuroQuery~\cite{neuroquery}
    & 4.266$\pm$0.262 & 7.560$\pm$0.389 & 36.055$\pm$0.761 & - & - & - \\

    & NeuroConText~\cite{neurocontext2}
    & 9.964$\pm$0.663 & 15.033$\pm$0.808 & 45.120$\pm$1.098 & 10.046$\pm$0.683 & 15.183$\pm$0.790 & 45.168$\pm$1.077\\

    & \cellcolor{gray!20}MNM-Reverse
    & \cellcolor{gray!20}8.852$\pm$1.224
    & \cellcolor{gray!20}13.486$\pm$1.342 & \cellcolor{gray!20}42.585$\pm$3.000 & \cellcolor{gray!20}10.332$\pm$0.562 & \cellcolor{gray!20}15.096$\pm$0.964 & \cellcolor{gray!20}44.636$\pm$1.270 \\
    
    & \cellcolor{gray!20}MNM $w/o\enspace\mathcal{L}_\mathrm{hier}$
    & \cellcolor{gray!20}10.699$\pm$0.650
    & \cellcolor{gray!20}15.807$\pm$0.653 & \cellcolor{gray!20}45.734$\pm$1.286 
    & \cellcolor{gray!20}10.458$\pm$0.590 & \cellcolor{gray!20}16.218$\pm$0.764 & \cellcolor{gray!20}46.406$\pm$1.057 \\

    & \cellcolor{gray!20}MNM
    & \cellcolor{gray!20}\textbf{10.728$\pm$0.673}
    & \cellcolor{gray!20}\textbf{15.860$\pm$0.668} 
    & \cellcolor{gray!20}\textbf{45.864$\pm$1.417}
    & \cellcolor{gray!20}\textbf{10.593$\pm$0.627}
    & \cellcolor{gray!20}\textbf{16.320$\pm$0.822} 
    & \cellcolor{gray!20}\textbf{46.590$\pm$1.217}\\
    \Xhline{2.5\arrayrulewidth}

    \multirow{6}{*}{\rotatebox[origin=c]{90}{Body}} & Text2Brain~\cite{text2brain}
    & 0.237$\pm$0.014 & 0.479$\pm$0.015 & 4.832$\pm$0.079 & - & - & - \\
    & NeuroQuery~\cite{neuroquery}
    & 4.421$\pm$0.374 & 7.696$\pm$0.458 & 36.055$\pm$0.562 & - & - & - \\
    & NeuroConText~\cite{neurocontext2}
    & 13.660$\pm$0.520& 20.228$\pm$0.742 & 51.945$\pm$0.961 & 13.747$\pm$0.579 & 19.517$\pm$0.449 & 51.572$\pm$0.771 \\

    & \cellcolor{gray!20}MNM-Reverse
    & \cellcolor{gray!20}13.756$\pm$0.575
    & \cellcolor{gray!20}20.136$\pm$0.485 & \cellcolor{gray!20}51.969$\pm$0.736 & \cellcolor{gray!20}14.087$\pm$0.556 & \cellcolor{gray!20}21.176$\pm$0.575 & \cellcolor{gray!20}53.134$\pm$0.845 \\
    
    & \cellcolor{gray!20}MNM $w/o\enspace\mathcal{L}_\mathrm{hier}$
    & \cellcolor{gray!20}14.288$\pm$0.747
    & \cellcolor{gray!20}20.649$\pm$0.850 & \cellcolor{gray!20}52.670$\pm$0.735 & \cellcolor{gray!20}14.569$\pm$0.648 & \cellcolor{gray!20}21.302$\pm$0.396 & \cellcolor{gray!20}53.342$\pm$0.797 \\

    & \cellcolor{gray!20}MNM
    & \cellcolor{gray!20}\textbf{14.434$\pm$0.618}
    & \cellcolor{gray!20}\textbf{20.741$\pm$0.574} 
    & \cellcolor{gray!20}\textbf{52.709$\pm$0.928}
    & \cellcolor{gray!20}\textbf{14.806$\pm$0.449}
    & \cellcolor{gray!20}\textbf{21.360$\pm$0.414}
    & \cellcolor{gray!20}\textbf{53.623$\pm$0.698}\\
    
    \Xhline{3\arrayrulewidth}
\end{tabular}}
\label{tab:classification}
\end{table*}

\section{Results and Analysis}

\subsubsection{Brain-Text Cross Modal Retrieval.}
\label{sec:cross_retireval}
To evaluate our semantic alignment performance, 
we perform 
text-to-brain and brain-to-text retrievals.
In the text-to-brain setting, the model ranks candidate brain activations 
by their relevance to given textual descriptions (e.g., abstracts or full-body text).
The brain-to-text setting reverses it, 
from a 
brain activation
to the most relevant article.
We adopt a 10-fold cross-validation 
to rigorously evaluate recall@K, $K\in\{5,10,100\}$.
Recall@K 
evaluates whether 
the true pair appears among the top-K 
similar candidates,
making it suitable for ranking tasks.
For comparison, the following baselines are adopted. 
Since Text2Brain~\cite{text2brain} and NeuroQuery~\cite{neuroquery} directly generate 3D brain activations 
from text, 
mean squared error (MSE) is used as similarity measure between the predicted brain activation and each candidate.
Unlike \cite{neurocontext2} using cosine similarity,
MNM adopts the negation of the absolute exterior angle.

For MNM, we employ two additional settings: (1) MNM-Reverse that reverses the hierarchy of MNM by flipping the brain and text relationship in $\mathcal{L}_\mathrm{ext}$ and $\mathcal{L}_\mathrm{cent}$, 
and (2) MNM $w/o\enspace\mathcal{L}_\mathrm{hier}$. 
As shown in Table~\ref{tab:classification}, our method consistently outperforms all baselines in both text-to-brain and brain-to-text retrieval, 
with the largest improvement observed in brain-to-text retrieval when using full-body text.
Degraded performances in two additional settings demonstrate the rationality of brain-text hierarchy and the need for explicit hierarchy guidance. 
 
\begin{figure}[t!]
\centering
\includegraphics[width=0.9\textwidth
]{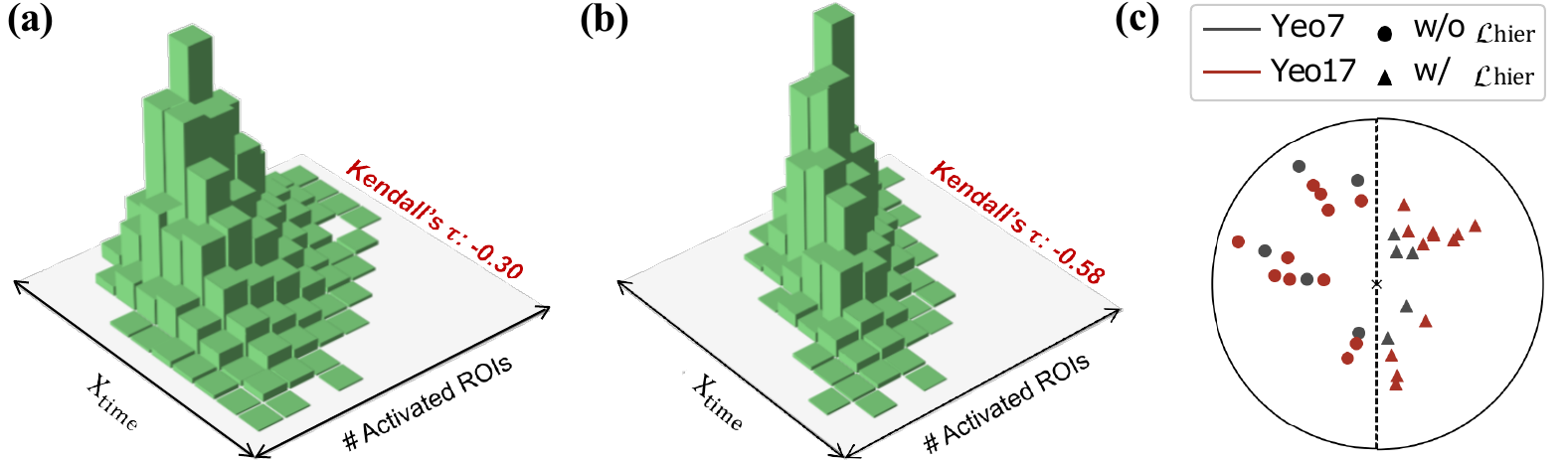}\\
    \caption{\footnotesize 
    \textbf{Effect of Brain Hierarchical Loss.}  
    (a) Histogram between $x_\mathrm{time}$ and the number of activated ROIs without $\mathcal{L}_\mathrm{hier}$, showing $-0.30$ of Kendall's $\tau$~\cite{kendalltau}  
    (b) Same histogram with $\mathcal{L}_\mathrm{hier}$, decreasing Kendall's $\tau$ to $-0.58$.  
    (c) 2D Poincar\'e disk projection of Yeo7 and Yeo17 network~\cite{yeo} embeddings (left: w/o $\mathcal{L}_\mathrm{hier}$, right: w/ $\mathcal{L}_\mathrm{hier}$).  
    }
    \label{fig:kendal_tau}
\end{figure}

\subsubsection{Brain Hierarchy Analysis.}
\label{sec:brain_hierarchy_analysis}
To show the effect of 
$\mathcal{L}_\mathrm{hier}$,
we analyze how samples are distributed in the latent space.
One of our 
goals is to model the inherent hierarchy of brain activation patterns,
such that more general (i.e., widely activated) regions lie closer to the hyperbolic origin than more specific (i.e., smaller or functionally narrower) ones.
To verify 
the assumption 
with the trained embeddings using 
\(\mathcal{L}_{\mathrm{hier}} \), 
we plot histograms of \( x_\mathrm{time} \) against the number of activated ROIs,
along with 
Kendall’s $\tau$~\cite{kendalltau} to quantify the ordinal correlation 
in a range $\left[-1,1\right]$.
A negative \(\tau\) indicates that embeddings with 
broader activations 
tend to reside at a higher ``level'' (i.e., smaller \( x_\mathrm{time} \)) in the hyperbolic space.
In Fig.~\ref{fig:kendal_tau},
while embeddings are dispersed (a) without \(\mathcal{L}_{\mathrm{hier}} \) resulting in -0.30 of Kendall’s $\tau$,
they 
are more inversely aligned 
(b) with \(\mathcal{L}_{\mathrm{hier}} \) resulting in Kendall’s $\tau=-0.58$.

We further examine two well-known cortical parcellation schemes, 
Yeo7 and Yeo17 networks~\cite{yeo}.
The Yeo7 network segments the brain into seven broad functional systems, 
while the Yeo17 network refines this 
into 17 detailed subdivisions.
To visualize these divisions, 
we map them into the learned hyperbolic space and project them onto a 2D Poincar\'e disk~\cite{poincare_disk}.
Due to page limitations, we visualize only half of the Poincar\'e disk,
varying the presence of $\mathcal{L}_{\mathrm{hier}}$.
In Fig.~\ref{fig:kendal_tau}(c), 
the embeddings from 
both networks exhibit no discernible order in the left half when $\mathcal{L}_{\mathrm{hier}}$ is absent.
However, with $\mathcal{L}_{\mathrm{hier}}$, 
Yeo7 embeddings cluster near the origin in the right half, enhancing interpretability.
This organization supports a coarse-to-fine analysis by selecting $x_\mathrm{time}$ from brain activation map candidates.


\begin{figure}[t!]
\centering
\includegraphics[width=0.98\textwidth]{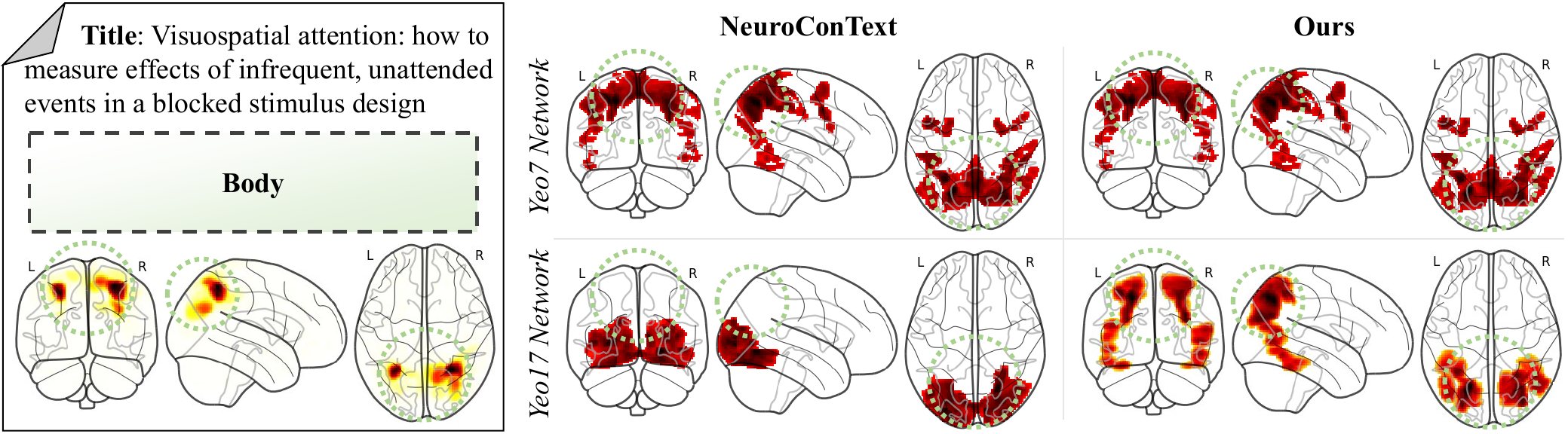}\\
    \caption{\footnotesize 
    \textbf{Reconstructed Brain Activation Map from a Neuroscience Article~\cite{visuospatial}.}  
    While NeuroConText~\cite{neurocontext2} fails to capture activation in the intraparietal sulcus using Yeo17 networks, MNM 
    assigns high similarity scores to 
    them as in the ground truth.  
    }
    \label{fig:text2brain}
\end{figure}

\subsubsection{Brain Activation Map Prediction.}
\label{sec:brain_prediction}
To evaluate text-brain alignment
beyond the task-specific decoder,
we use the embeddings of Yeo7 and Yeo17 networks~\cite{yeo} as our brain activation bases.
With body text of given articles,
we calculate the similarity of the text embedding 
and brain activation bases of two networks.
NeuroConText~\cite{neurocontext2} 
is a unique baseline, 
since it aligns both modalities in the shared latent space as we do. 
We employ the same similarity metric as in the brain-text cross-modal retrieval experiment.
The resulting similarities are transformed into 
probabilities 
using a softmax function 
and assigned to the corresponding voxels in the 3D brain.
For better visualization,
we threshold the top 10\% of voxels and visualize 
using brain glass schematics via the Nilearn~\cite{nilearn} software.

Fig.~\ref{fig:text2brain} illustrates how embeddings can be recovered using the Yeo7 and Yeo17 networks for a study~\cite{visuospatial} that investigates the neural mechanisms underlying visuospatial attention reorientation.
The study demonstrates that both blocked and event-related analyses consistently activate the intraparietal sulcus and superior parietal cortex.
In coarse prediction using the Yeo7 networks, both NeuroConText and MNM exhibit a high correlation (i.e., top 10\% similarity) with the activated regions. 
However, with the finer-grained Yeo17 network,
NeuroConText fails to capture the target regions, whereas our method successfully identifies them.
Since most detected regions overlap across different parcellation levels, 
MNM demonstrates greater consistency compared to NeuroConText. 
These results indicate that our hyperbolic representation can flexibly adapt to various parcellation schemes when inferring activation maps from articles.

\section{Conclusion}

In this work, we propose MNM, a novel framework that leverages hyperbolic geometry to capture the inherent hierarchical structure of brain activations 
while aligning them more effectively with corresponding textual descriptions.
Through extensive experiments, we demonstrate that MNM not only surpasses baselines in cross-modal retrieval but also provides greater interpretability in activation map prediction.
By providing robust performances across various parcellation schemes,
MNM paves the way for multi-level neuroimaging meta-analysis.

%
%

\begin{credits}
\subsubsection{\ackname} 
This research was supported by NRF-2022R1A2C2092336 (50\%), RS-2025-02216257 (20\%), RS-2022-II220290 (20\%), and RS-2019-II191906 (AI Graduate Program at POSTECH, 10\%).

\subsubsection{\discintname}
The authors have no competing interests to declare that are
relevant to the content of this article.

\end{credits}

\bibliographystyle{splncs04}
\bibliography{Paper-0365}




\end{document}